\documentclass{svproc}
\usepackage{url}
\usepackage{color}
\usepackage{romannum}
\usepackage{caption}
\usepackage[numbers]{natbib}
\usepackage{algorithm}
\usepackage{algpseudocode}
 \usepackage{amsmath}
\usepackage{amssymb}
\usepackage{amsfonts}
\usepackage{graphicx}
\usepackage{multirow}
\usepackage[utf8]{inputenc}
\usepackage{float}
\let\oldref\ref
\renewcommand{\ref}[1]{(\oldref{#1})}
\makeatletter
\def\algbackskip{\hskip-\ALG@thistlm}
\makeatother
\algnewcommand\algorithmicforeach{\textbf{for each}}
\algdef{S}[FOR]{ForEach}[1]{\algorithmicforeach\ #1\ \algorithmicdo}

\begin{document}
\mainmatter              % start of a contribution
\title{GMC: Grid Based Motion Clustering in Dynamic Environment}
\titlerunning{GMC in dynamic environment}  % abbreviated title (for running head)
% \inst{\Letter}
\author{Handuo Zhang \and Karunasekera Hasith \and Han Wang}
\authorrunning{Handuo Zhang et al.} % abbreviated author list (for running head)
\tocauthor{Handuo Zhang, Karunasekera Hasith, and Han Wang}
\institute{School of Electrical \& Electronic Engineering,\\ Nanyang Technological University, Singapore,\\
\email{ \{hzhang032, karu0009\}@e.ntu.edu.sg, hw@ntu.edu.sg } %, 
%\\ home page:\texttt{http://zhanghanduo.github.io}
}

\maketitle              % typeset the title of the contribution

\begin{abstract}
Conventional SLAM algorithms takes a strong assumption of scene motionlessness, which limits the application in real environments. 
This paper tries to tackle the challenging visual SLAM issue of moving objects in dynamic environments. We present GMC, grid-based motion clustering approach, a lightweight dynamic object filtering method that is free from high-power and expensive processors. GMC encapsulates motion consistency as the statistical likelihood of detected key points within a certain region. 
Using this method can we provide real-time and robust correspondence algorithm that can differentiate dynamic objects with static backgrounds.

We evaluate our system in public TUM dataset. To compare with the state-of-the-art methods, our system can provide more accurate results by detecting dynamic objects.
% We would like to encourage you to list your keywords within
% the abstract section using the \keywords{...} command.
\keywords{visual SLAM, motion coherence, dynamic envrionment} %multiview geometry
\end{abstract}
\section{Introduction}
Localization has always been a core and challenging task in mobile robotics, autonomous driving, autonomous UAVs and other navigation applications. With the development of all kinds of smart algorithms together with more powerful sensors and processors, now the translation error of SLAM has even been reduced to around $0.8\%\sim 1.2\%$ according to KITTI odometry and SLAM evaluation benchmark~\cite{Geiger2012CVPR}\footnote{KITTI Benchmark: http://www.cvlibs.net/datasets/kitti/eval\_odometry.php} with monocular, stereo  and RGBD cameras. Some significant visual odometry methods like SVO~\cite{forster2014svo, forster2017svo}, ORB-SLAM~\cite{mur2016orb}, DSO-SLAM~\cite{engel2018direct} and S-PTAM~\cite{pire2017s}, etc are listed in the rankings. They are mostly inspired by the groundbreaking key-frame based monocular SLAM framework PTAM~\cite{klein2007parallel} and become popular partially because cameras are relatively cheap and handy to set up, and they passively acquire texture-rich cues from the environment, which makes it applicable in real-world environments.

However, most of current visual SLAM system are still sensitive to highly dynamic scenarios due to the strong assumption that the surroundings are static. Currently unexpected change of surroundings will probably corrupt the quality of the state estimation process and even lead to system corruption. For example, the presence of dynamics in the environments like walking pedestrians and moving vehicles, might give rise to misleading data association for vision based SLAM systems. Although there have been some researches to reduce the sensitivity to dynamic objects on laser based SLAM~\cite{wolf2005mobile, rodriguez2007improved} based on Iterative Closest Point (ICP) method, the problem is still not well studied on vision-based SLAM. 

In recent years with the rapid development of deep learning, some researchers turn to high-level understanding of the surroundings~\cite{ren2017faster, liu2016ssd, redmon2016you}. This can help remove the moving objects and indirectly solve the problem. However, the property of deep neural network makes the hardware requirements really high, also real-time performance can hardly be reached on mobile robot platforms.

In this paper, we focus on reducing the impact of dynamic objects without the need of neural network, which can be employed on lightweight mobile platform. The main contributions of our work are:

\begin{itemize}
    \item We proposed a dynamic feature point filtering method called grid-based motion clustering (GMC). It can significantly reduce the impact of dynamic object on pose estimation.
    \item The GMC method also can used to reject outliers during feature matching based on motion consistency within certain neighborhood.
    \item We build a complete SLAM system that adopts GMC method and the performance of the system is validated on TUM dataset. 
    \item the proposed approach can be easily integrated into other existing SLAM frameworks.
\end{itemize}

The rest of this paper is organized as follows. Related work is presented in Sec.~\ref{sec.related}. Sec.~\ref{sec.GMC} discusses the proposed algorithm, and based on that we discuss how to integrate into a SLAM system. Finally in Sec.~\ref{sec.exp} we evaluate the performance of our proposed framework in the popular KITTI benchmark and in real applications, validating our performance.

%Some paper on deep learning based methods. Object detection; segmentation based mask.

\section{Related Work} \label{sec.related}
The basic assumption of most SLAM methods is that the landmarks observed are static, which limits the application of visual SLAM. Some recent researches start researching active moving objects using dense SLAM by incorporating optical flow technique. Optical flow is generated with movement of pixels, so theoretically static background and moving targets can be distinguished by that ~\cite{alcantarilla2012combining, tan2013robust}. However, this kind of dense sampling requires a large amount of calculation. Also as the association cannot be guaranteed, these methods are not robust, and RANSAC might fail easily, especially when the object size is big. Some learning based methods are utilized to minimize the reprojection error or pixel intensity with the association issue one of the values to be solve using EM~\cite{bowman2017probabilistic} or CRF~\cite{tipaldi2009motion}, but the results cannot be achieved in real-time.

Semantic segmentation has gained significant improved over past year due to deep learning based methods. Segnet \cite{badrinarayanan} and Mask R-CNN \cite{he2017mask} are two recent popular semantic segmentation methods. Semantic segmentation provides the information for which object (e.g. human, car, road, building and etc.) each pixel in the image belongs to. In SLAM problem features are tracked across frames to localize the camera. However, these features suffer from illumination variance and vie-point variations making them loose track in few frames. Furthermore, most SLAM approaches assume that the most parts of the image represent the static environment, which is not the ideal assumption for a reliable road scenes. Semantic segmentation provides the high level scene understanding which could facilitate rectifying the errors introduced in the above two scenarios.

Simply associating semantic information with multiple fixed points VO can be improved \cite{lianos2018vso} because of the long term association of features across frames. Other than point features, object level information can also be used in localizing ego position, coupled with inertial and point features into a single optimization framework \cite{bowman2017probabilistic}. In extreme viewing variations due to occlusion, large illumination changes based on the time of the day and viewing angle changes can cause great challenges to localization problem. \cite{schonberger2018semantic} proposes a method to handle this problem using a generative model to learn descriptors combining geometric and semantic information of the scene.  Additionally semantic information can be used to classify static and dynamic objects and thus features belonging to static and dynamic classes. DS-SLAM \cite{yu2018ds}, DynaSLAM \cite{bescos2018dynslam} and SIVO \cite{ganti2018visual} uses semantic information to remove the outliers in consistently moving features in dynamic objects such as humans. DynaSLAM \cite{bescos2018dynslam} later uses semantic info to inpaint the occluded static scenes. In contrast to just using semantic information to remove dynamic objects, it can be used to calculate both ego motion and motions of each object using stereo visuals \cite{li2018stereo}. Use of semantic information is useful to keep only the static permanent information (e.g. features on buildings, road and etc.) when generating the map \cite{yu2018ds} while making use of static temporary information (e.g. features on parked vehicles) and dynamic information (e.g. features on dynamic vehicles and persons) in local frames to improve the localization.

\section{Grid based Motion Clustering Approach} \label{sec.GMC}
In some researches, CRF is utilized to do clustering considering the distance between pairwise point distance and neighboring constraints~\cite{tipaldi2009motion}. This type of approach derives from the idea of image segmentation and cannot be achieved in real-time. In this section the GMC is addressed for dynamic object clustering even without counting every pixel. The strategy of accumulating keypoints and classification based distance measurement makes the process fast. Meanwhile, the Quadtree~\cite{samet1984quadtree} aided adjustment process makes the algorithm robust.

This approach utilizes the basic property of motion coherence constraints, i.e. neighboring pixels share similar motion. While in localization task, exhaustively comparing each pixel is time expensive. So we take discriminative features considering motion coherence as a statistics likelihood within a region. This property will be proved in section~\ref{subsec:coherence}.

In the following, we will extract the feature points and explicitly use the initial guess of rotational $R$ and translational $T$ parameter of the motion vector to convert the triangulated new frame keypoints into 3D locations in global coordinate, so as to further calculate the distance measurements which will be discussed in sec.~\ref{subsec:distance}.

% approximate size of each grid is $N/G$ points. However, the distribution of the most discriminative keypoints is not uniform. So we have to guess the "center" of that grid based on the distribution of features at that frame. Our strategy is to select the keypoint that has the highest score and dynamically adjust this center iteratively according to the distance measurement.

%\subsection{Notation}
%
%Image pairs.

\subsection{Distance Measurement}\label{subsec:distance}

The normal metrics of 3D points is the Euclidean distance between them in the reference frame $\mathcal{F}_{re}$ and the associated frame $\mathcal{F}_{ma}$.

\begin{equation}\label{eq.ba}
% e = \sum_{j=1}^{M} \sum_{s_i \in{L_j}} \| r_i - ( T_j + R_j m_i )  \|^2
e_i^j = \| re_i - ( T_j + R_j ma_i )  \|^2
\end{equation}
where $j$ represents the frame index, $i$ is the feature point index, thus $R_j$ and $T_j$ are initial estimated rotation and translation parameters. 

For an image with $M$ keypoints which are all static, the theoretical assumption is that their 3D displacement of keypoint locations are zero. While for dynamic objects, displacement does not infer full motion information, i.e. does not differentiate any motion directions, like illustrated in Fig.~\ref{fig:motion}. Unlike 2D reprojection error, the 3D Euclidean error triangulated from 2D feature point pairs are impacted by the distance. So on top of equation~\ref{eq.ba} we incorporate the covariance of reconstructed 3D scene feature points ~\cite{beder2006determining}. According to multiview geometry, we can conclude that the covariance of 3D reconstructed points have relationship with 2D image point covariance matrices, the 3D locations, and the projection matrices. 

\begin{eqnarray}
cov = 
\left(
\begin{array}{cc}
\mathbf{A}^{\intercal} \left( \mathbf{B}  \left(
	\begin{array}{cc}
	\mathbf{C}_1 & 0 \\
	0 & \mathbf{C}_2 \\
	\end{array} \right) \mathbf{B}^{\intercal}  \right)^{-1} \mathbf{A}

 \end{array}\right)
\end{eqnarray}
where $\mathbf{C}^1$ and $\mathbf{C}^2$ are image coordinate covariance matrices respectively, $\mathbf{A}$ is defined as the inverted 3D location, $\mathbf{B}$ defined as the inverted 2D point location. To simplify, we add a parameter to normalize the 3D residuals which is proportional to the inverse 3D distance $\mathbf{X}$:

\begin{equation}\label{eq.dis}
e_i^j = (\mathbf{X}^{re}_i - ( T_j + R_j \mathbf{X}^{ma}_i )) \cdot \alpha / (\mathbf{X}^{ma})^{-1}
\end{equation}
where $\alpha$ is a parameter to constraint the geometry residuals inside a certain scope for next step processing.

Considering both discrimination and calculation efficiency, we choose motion classification, with finite combinations of motion clusters in $z$ direction and $x$ direction respectively. The merits of this proposed distance measurement is for the further motion statistics and clustering convenience sake, with any motion residuals lying within a predefined intervals counted as this motion pattern.

We set the minimal geometry residual $e_{int}^z$, $e_{int}^x$ as the unit of the finite motion category pattern. Then the distance measurement can be converted into a 2D motion pattern table with $e_{int}$ the interval, as shown in Fig~\ref{fig:motion}.

\noindent%
\begin{minipage}{\linewidth}% to keep image and caption on one page
    \vspace{1ex}
    \makebox[\linewidth]{%        to center the image
        \includegraphics[width=1.0\linewidth]{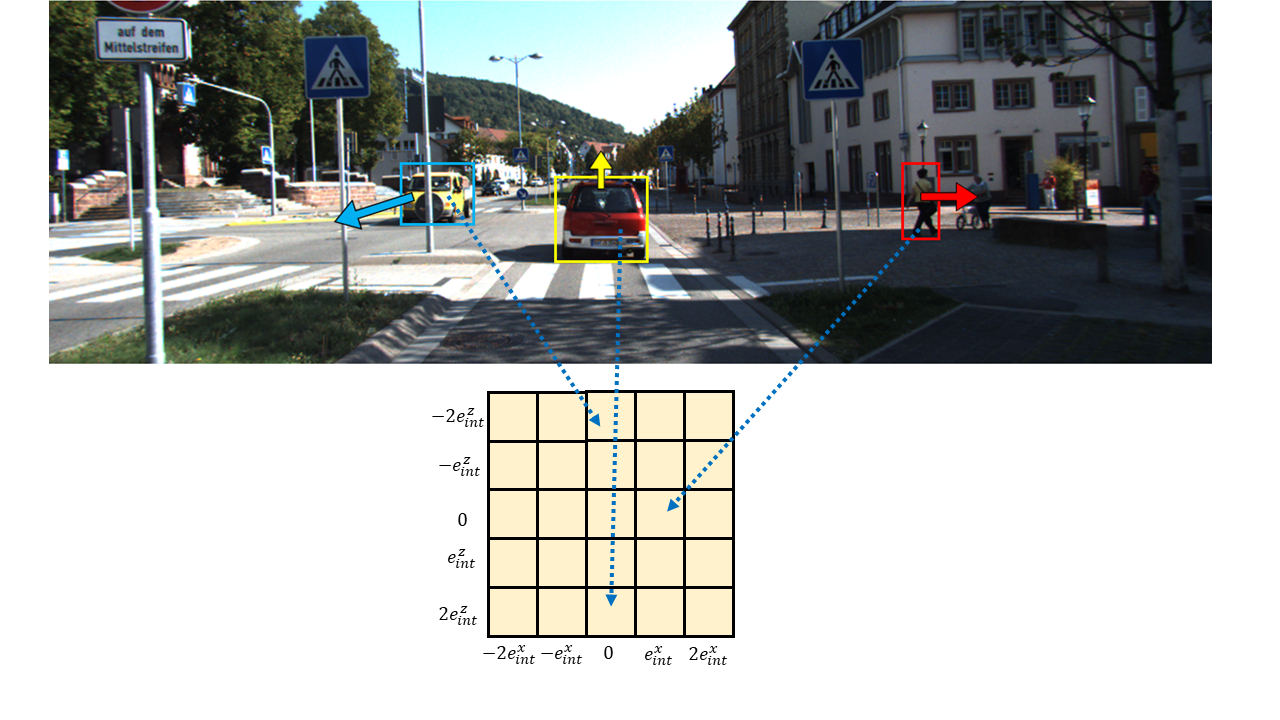}}
    \captionof{figure}{Illustration of dynamic object motion patterns. The center cell represents static status.}\label{fig:motion}%      only if needed  
\end{minipage}

\subsection{Motion Constraint Analysis} \label{subsec:coherence}

Given adjacent frame pairs, the associated extracted feature points should view the same 3D locations. So its neighboring keypoints shall move together if the density of features is high enough.

According to the conclusion of \cite{bian2017gms}, motion smoothness causes a small neighborhood around a true match to view the same 3D location. Likewise, it also causes a (small) neighborhood around a object keypoint to drop into the same motion pattern table as shown in~\ref{fig:motion}, which is also the primitive idea of the proposed approach. $re$ refers to reference frame, $ma$ refers to matched frame. We set $f_{re}$ as one of the $n$ features in region $re$. Thus $f_{re}^{ma}$ represents $f_{re}$'s nearest neighbor happens to be the feature in region $ma$.

Therefore our assumption gives

\begin{equation} \label{eqn.smoothness}
    p(f^{ma}_{re} | f_{re}^{false}) = \beta m/M
\end{equation}
where $M$ is all the location possibilities once the $f_{re}$'s neighbors drop into different local regions from this keypoint, $m$ is the number of features in region $ma$ and $\beta$ is a factor to accommodate violations of this assumption caused by repeated structures like a row of windows. 

Let $p_{true} = p(f^{ma}_{re} | T^{re\rightarrow ma})$ be the probability that, given ${re, ma}$ view the same location, feature $f_{re}$'s nearest neighbor is in region $ma$. Also we set the event of motion consistency with $T^{re\rightarrow ma}$ the same motion of all keypoints in one local region and $F^{re\rightarrow ma}$ not the same motion.  So we can derive:

\begin{align} \label{eq.pt}
\begin{split}
p_{true} =  p(f^{ma}_{re} | T^{re\rightarrow ma}) 
& = p(f^{true}_{re} | T^{re\rightarrow ma}) + p(f^{false}_{re}, f^{ma}_{re}| T^{re\rightarrow ma}) \\
& = p(f^{true}_{re} | T^{re\rightarrow ma}) + \\ & p(f^{false}_{re}|T^{re\rightarrow ma})p(f^{ma}_{re}|f^{false}_{re} , T^{re\rightarrow ma}) \\
& = t + (1-t)\beta m / M \\
\end{split}
\end{align}
where $t$ represents the probability of all $f_{re}$'s neighbors are consistent with $f_{re}$.

Likewise, let $p_{false} = p(f^{ma}_{re} | F^{re\rightarrow ma})$ and similar to \ref{eq.pt}, 

\begin{align} \label{eq.pf}
p_{false} = \beta(1-t) m / M 
\end{align}

According to equations \ref{eq.pt}, \ref{eq.pf} we can approximate the distribution of $S_i$, the number of points in a neighborhood of keypoint $x_i$, with a pair of binomial distribution:

\begin{equation}
S_i \sim 
\begin{cases}
    B(n, p_{true}), & \text{if } x_i \text{ is true} \\
    B(n, p_{false}), & \text{if } x_i \text{ is false} 
\end{cases}
\end{equation}
where $S_i$ here refers to the neighborhood scores that support the motion of the target keypoint $k_i$. The two distinct distributions make $S$ the good indicator for considering a local region with evident motions different from background. In the next section, we will use this property to solve dynamic point filtering by counting the neighborhood scores and merge similar motion patterns.

\subsection{Approach Description} \label{subsec:descript}

The proposed algorithm takes a presupposed number of approximately equally sized grid cells $N$. Thus a motion statistics tensor is created as shown in Fig~\ref{fig:tensor}~(a).

\subsubsection{Assignment with Consensus of motion patterns}
\hfill \break

Enough number of keypoints can provide statistically better quality motion cluster. Referring to Fig~\ref{fig:tensor}, each layer of the tensor represents the motion distribution of the associated grid cell. So we can just calculate the residuals of the location difference $\mathbf{x}_j^{ma} - \mathbf{x}_j^{ref}$ where $j$ is the keypoint index within this grid and $\mathbf{x} \in{\mathbb{R}^3}$.

\noindent%
\begin{minipage}{\linewidth}% to keep image and caption on one page
    % \vspace{1ex}
    \makebox[\linewidth]{%        to center the image
        \includegraphics[width=1.0\linewidth]{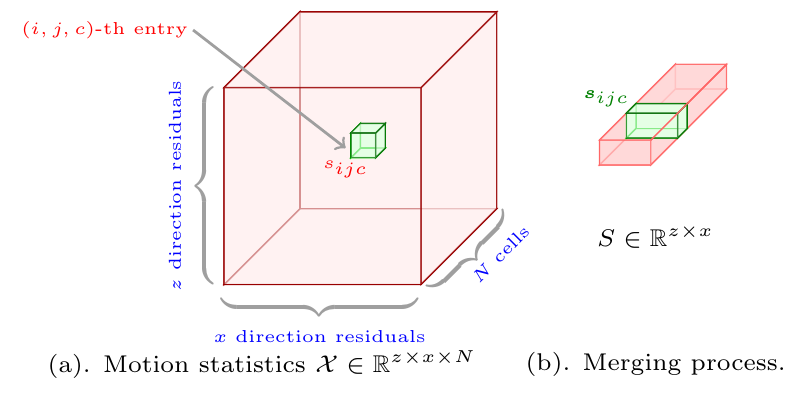}}
    \captionof{figure}{Fig. (a) illustrates the motion clustering tensor $\mathcal{X}\in{\mathbb{R}^{z\times x \times N}}$ where $N$ is the number of grid cells. We classify the dynamic objects into 2D arrays $S\in{\mathbb{R}^{z \times x}}$ for $z$ and $x$ directions respectively. 
        Fig. (b) demonstrates the merging process between all the cells that share similar motions.} \label{fig:tensor}
\end{minipage}

\subsubsection{Assignment with Quadtree}
\hfill \break

The counting of grid cells with the most feature points falling in sounds easy but will cause the motion segmentation not accurate due to the fixed size of grid cell, which in some cases, may lead to failure by not considering the partial object cased in the cell. So hereby we proposed quadtree structure to further split the grid cell into quads and repeat the assignment task.

The condition of subdividing grids is twofold:
\begin{enumerate}
    \item The motion pattern that wins the consensus has fewer than $p_{min} \cdot{n_i}$, with $p_{min}$ the threshold of maximum point proportion and $n_i$ the number of all keypoints detected within cell $i$.
    \item $n_i > n_{min}$ with $n_{min}$ the minimum  required keypoints for statistics. Our motion smoothness is valid only when there are enough number of points, as illustrated in equation~\ref{eqn.smoothness}.

\end{enumerate}

This subdividing process ends when one of the above conditions fails. Here we display the pseudo-code for the whole procedure of assignment.

\begin{algorithm}[H]
    \caption{GMC Assignment}\label{algo:gmc_assignment}
    \hspace*{\algorithmicindent} \textbf{Input} $\mathcal{F}_r$:reference frame, $\mathcal{F}_m$: frame to be matched, $\mathbf{R}_0,\mathbf{t}_0$: initial estimation of pose transform.\\
    \hspace*{\algorithmicindent} \textbf{Output} $\mathcal{L}$: motion pattern tensor representing most likely motion clusters for each sub-region of the frame.
    \begin{algorithmic}[1]
        
        %step 1
        %		\algorithmiccomment{\textbf{Step1}}\\
        
        \State Detect feature points with corresponding descriptors from $\mathcal{F}_r$ and $\mathcal{F}_m$: $\mathbf{s}_r$ and $\mathbf{s}_m$
        \State Calculate the 3D location residuals of the feature points $\mathbf{e}$. 
        \State Generate an image-size 2D array $I$ with height $h$ and width $w$. The indexes are the corresponding 2D key point coordinates $s_m$, and the values are the the 3D location residuals $\mathbf{e}$.
        \Comment{Eq. ~\ref{eq.dis}}
        \State Divide $I$ by $N = G_x \times G_y$ grid cells. Create a 3D tensor $\mathcal{X}$ with depth the grid number $N$, width $x$ and height $z$ the motion pattern resolution. \Comment{Fig. ~\ref{fig:motion}}
        
        \ForEach {keypoint $k\in{\mathcal{K}}$ in array $I$}
        \State Acquire the grid cell index $g$ this key point belongs to;
        \State push the 3D residuals $\mathbf{e}$ into the corresponding entry $s_{ijc}$ of $\mathcal{X}$. The count of this entry plus $1$.
        
        \EndFor
        
        \For {$c=1$ to $N$}
        \State Find the maximum value $s_{max1}$ and second biggest value $s_{max2}$ of the motion pattern $\mathcal{X}_c$.
        
        \EndFor
        
        \State Run motion clustering process as explained in Algorithm \ref{algo:merge}.
        \Repeat \\
        from line 4 to line 12, with motion patterns shifted by half cell-width in the $x$,$z$ and both $x$ and $z$ directions. (3 times)
        \Until The cluster of motion converges.
    \end{algorithmic}
    
\end{algorithm}

\subsubsection{Merging}

\hfill \break

After the assignment, we might have hundreds of motion entries. Thus an additional updating and merging operation is needed, because we need a post-processing step to enforce connectivity by reassigning disjoint pixels to nearby clusters. Here we have an assumption based on the prior knowledge that most regions of the background is static. So we hereby adopt a strong constraint: cluster the motion patterns as few as possible and impose more penalties to low motion areas to avoid trivial motion fragmentation. Thus we adopt non-maximum suppression approach to make sure there is no duplicate assignments in nearby neighborhoods.

We first eliminate the cluster with fewer than 10 features and sort the grids in each motion pattern and choose the highest score grids the candidates. Then we step by step merge clusters with the same label around the candidates.

%\begin{align}
%    \begin{split}
%    \log p(x|\mathcal{L}) = 
%    \end{split}
%\end{align}

%Prior knowledge that most clusters belonging to the "static" background that has no motion at all. So we add a regularization term $w$ to minimize the number of clusters.

The entire algorithm is summarized in Algorithm \ref{algo:merge}.

\begin{algorithm}
    \caption{Merge Clusters}\label{algo:merge}
    \hspace*{\algorithmicindent} \textbf{Input} $\mathcal{L}$: labels representing different motion patterns with the corresponding assignment. \\
    \hspace*{\algorithmicindent} \textbf{Output} New cluster assignment $\mathcal{L}\prime$. 
    \begin{algorithmic}[1]
        
        %step 1
        %		\algorithmiccomment{\textbf{Step1}}\\
        \Repeat
        \ForEach {motion pattern $m$} Sort all the grid cells within $m$ in decreasing order
            \ForEach {grid cell assignment $l_i \in{l_m}$} Find its neighbors with the same labels.
                \If {Can find} Merge $l_i$ and $l_k$ to create the new cluster $\mathcal{L}\prime$
                \EndIf
            \EndFor
        \EndFor
        \Until the distance between two cells is below threshold
        \State Return $\mathcal{L}\prime$
        
    \end{algorithmic}
    
\end{algorithm}

\subsection{Integration into SLAM System}

\begin{figure}[!t]
    \centering
    \includegraphics[trim=0 1cm 5cm 0,width=6in]{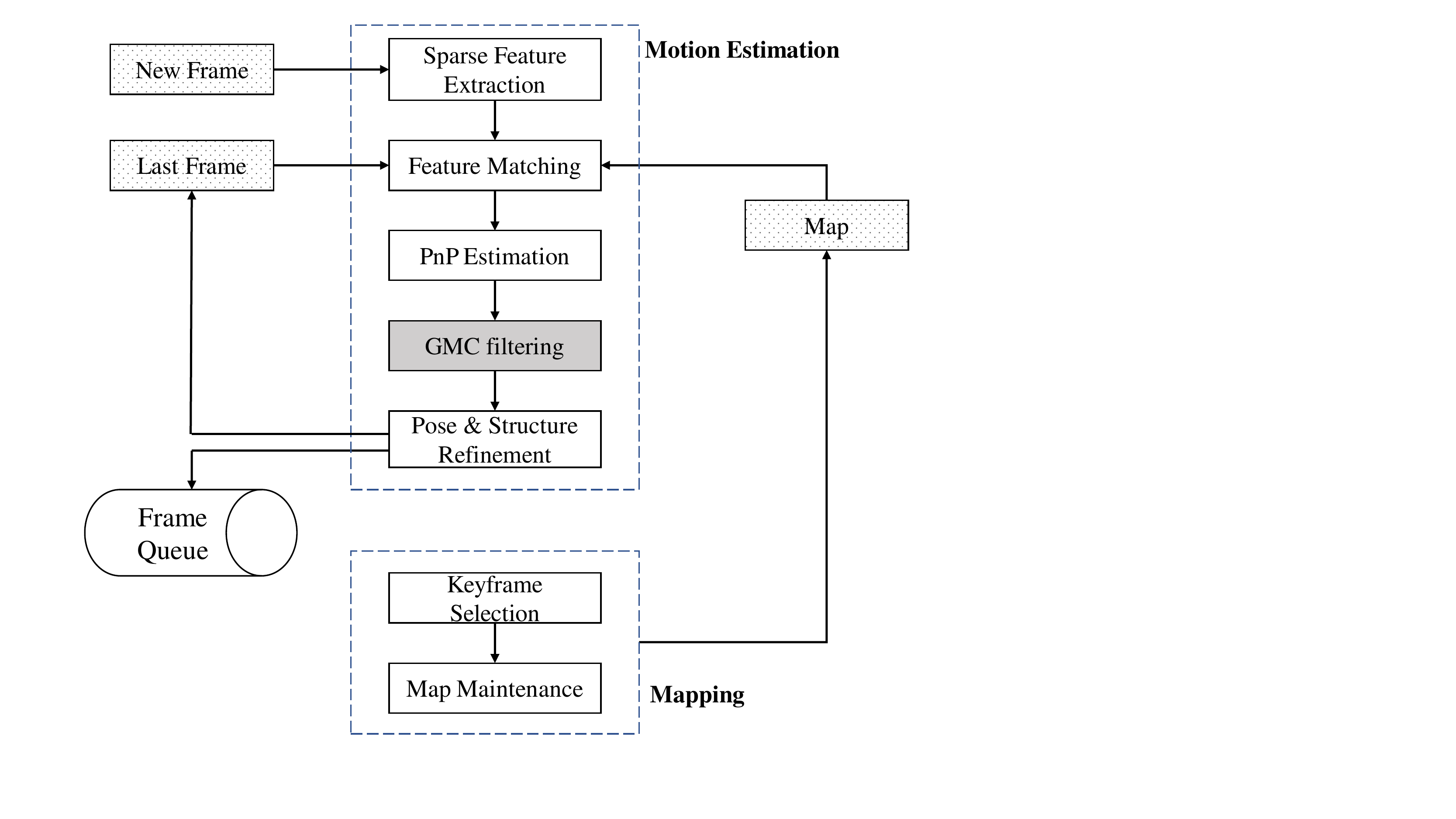}
    \caption{The pipeline of integrating GMC filtering approach into general visual SLAM system. }
    \label{fig:pipeline}
\end{figure}

Fig.\ref{fig:pipeline} illustrates an overview pipeline of the SLAM system. First of all, the new frame is passed into the system and motion estimation is operated as normal. The, before the bundle adjustment (pose and structure refinement) step, we insert the GMC filtering module which has been discussed in details in section~\ref{subsec:descript}.

\section{Experiments} \label{sec.exp}

In this section we present extensive experimental outcomes to
demonstrate the superior performance of the proposed system.
To benchmark the integration of GMC with SLAM methods, we inject GMC approach into ORB-SLAM2 (RGBD version) \cite{mur2016orb} and compare it with the original performance. The advantage of RGBD is that we don't need to triangulate the 3D points and calculate the covariance. However, we can still use monocular or stereo version of SLAM with GMC. The experimental results demonstrate the competitiveness of the proposed system as well as the versatile nature of the fusion framework.

All the experiments are performed on a standard laptop running Ubuntu 16.04 with 8G RAM and an Intel Core i7-6700HQ CPU at 2.60 GHz. 

\subsection{Filtering out dynamic points}

The input of our algorithm is the extracted features from two frames. As we use motion coherence advantage, based on equation~\ref{eq.pt}, \ref{eq.pf}, we need abundant information from features to correctly partition moving regions and static regions. In the experiment we use GFTT~\cite{shi1993good} features and set the nearest distance as 10 decrease the threshold to extract features. 

\begin{figure}[!t]
    \centering
    \includegraphics[width=1.0\linewidth]{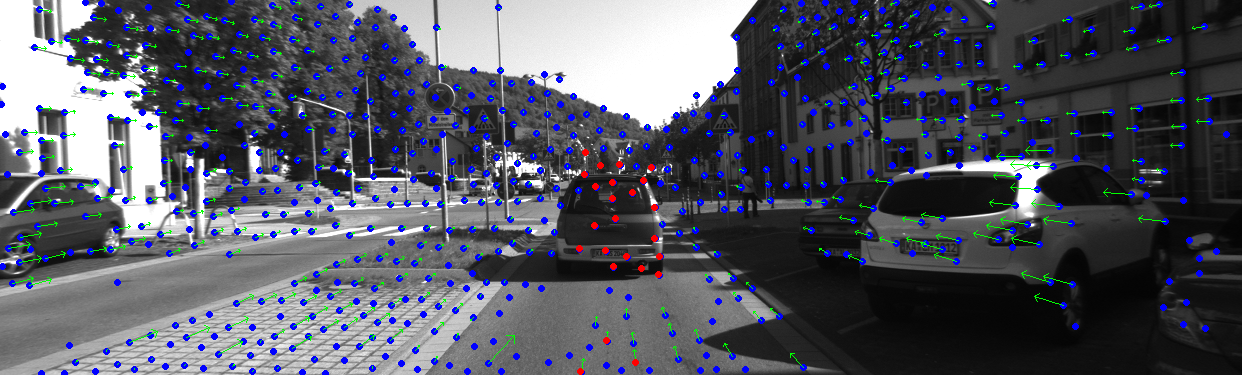}
    \includegraphics[width=1.0\linewidth]{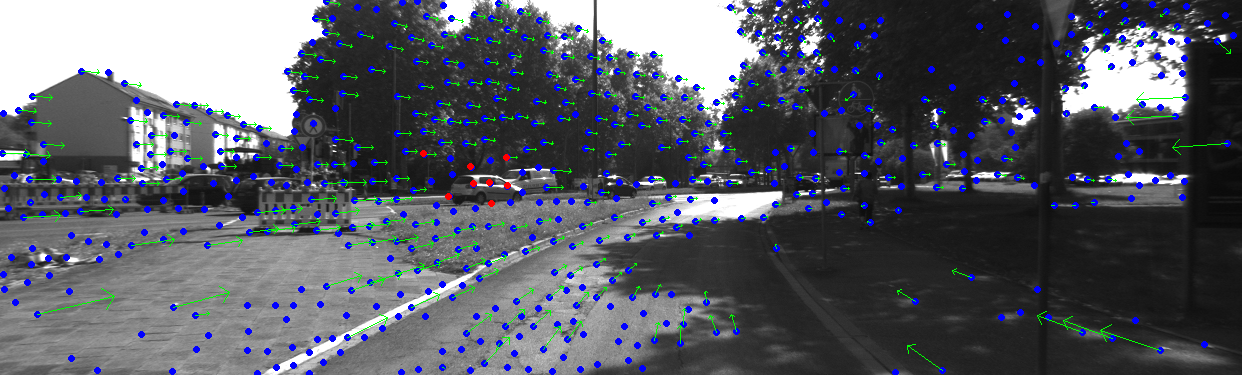}
    \caption{The demonstration of GMC filter on KITTI RAW dataset. Note that GMC can well handle dynamic object with certain size due to the grid resolution. To increase accuracy, you have to increase the number of features to be extracted. }
    \label{fig:filtering}
\end{figure}

For more features, RANSAC process will take much longer time and the result might be not robust. We adopt the grid based statistics matcher (GMS)~\cite{bian2017gms} to help reject outliers with proper matching time.
As shown in Fig~\ref{fig:filtering}, we evaluate our method in KITTI dataset with some middle traffic. It proves that our method can label the dynamic features with high accuracy. 

\subsection{Accuracy Evaluation}\label{subsec:accuracy}

We evaluate our results in TUM RGB-D dataset~\cite{sturm12iros} with dynamic scenes and accurate ground truth system capturing walking and sitting motions. In the experiment, we take the grid cell number $20 \times 15$ to balance the efficiency and accuracy.

\begin{table}[!t]
    \centering
    \caption{Comparison on translational error RMSE and MAE of ORBSLAM with and without GMC (ATE) }
    \begin{tabular}{l||c|c||c|c} 
        \hline\hline
        \multicolumn{1}{c||}{\multirow{2}{*}{Dataset}} &
        \multicolumn{2}{c||}{ORBSLAM+GMC} & \multicolumn{2}{c}{ORBSLAM} \\ 
        \cline{2-5} & RMSE & MAE & RMSE & MAE \\
        \hline
        fr3\_walking\_xyz & \textbf{0.0873} & \textbf{0.0451} & 0.8121 & 0.5977 \\
        fr3\_walking\_static & \textbf{0.0157} & \textbf{0.0193} & 0.4115 & 0.3198 \\ 
        fr3\_walking\_rpy & \textbf{0.7301} & \textbf{0.4935} & 0.8802 & 0.7992 \\
        fr3\_walking\_half & \textbf{0.1029} & \textbf{0.0644} & 0.5273 & 0.4100 \\
        fr3\_sitting\_static & 0.0088 & 0.0593 & 0.0089 & 0.0071  \\
%     \\\hline
%        \multicolumn{1}{c||}{mean} & \textbf{0.069} & \textbf{0.060} & 0.205 & 0.186 \\ 
        \hline \hline
    \end{tabular}
    \label{tab.ate}
\end{table}

\begin{table}[!t]
    \centering
    \caption{Comparison on translational error RMSE and MAE of ORBSLAM with and without GMC (RPE). (Unit: m) }
    \begin{tabular}{l||c|c||c|c} 
        \hline\hline
        \multicolumn{1}{c||}{\multirow{2}{*}{Dataset}} &
        \multicolumn{2}{c||}{ORBSLAM+GMC} & \multicolumn{2}{c}{ORBSLAM} \\ 
        \cline{2-5} & RMSE & MAE & RMSE & MAE \\
        \hline
        fr3\_walking\_xyz & \textbf{0.0977} & \textbf{0.0347} & 0.4344 & 0.2469 \\
        fr3\_walking\_static & \textbf{0.0137} & \textbf{0.0094} & 0.2120 & 0.0208 \\ 
        fr3\_walking\_rpy & \textbf{0.1943} & \textbf{0.0735} & 0.4501 & 0.1507 \\
        fr3\_walking\_half & \textbf{0.0338} & \textbf{0.0303} & 0.3441 & 0.0673 \\
        fr3\_sitting\_static & \textbf{0.0081} & \textbf{0.0066} & 0.0093 & 0.0070  \\
        %     \\\hline
        %        \multicolumn{1}{c||}{mean} & \textbf{0.069} & \textbf{0.060} & 0.205 & 0.186 \\ 
        \hline \hline
    \end{tabular}
    \label{tab.rpe}
\end{table}

\begin{table}[!t]
    \centering
    \caption{Comparison on rotational error RMSE and MAE of ORBSLAM with and without GMC (RPE) }
    \begin{tabular}{l||c|c||c|c} 
        \hline\hline
        \multicolumn{1}{c||}{\multirow{2}{*}{Dataset}} &
        \multicolumn{2}{c||}{ORBSLAM+GMC} & \multicolumn{2}{c}{ORBSLAM} \\ 
        \cline{2-5} & RMSE & MAE & RMSE & MAE \\
        \hline
        fr3\_walking\_xyz & \textbf{1.8765} & \textbf{0.9878} & 8.4355 & 6.7230 \\
        fr3\_walking\_static & \textbf{0.9818} & \textbf{0.2880} & 4.1566 & 0.6744 \\ 
        fr3\_walking\_rpy & \textbf{7.7448} & \textbf{2.1677} & 8.0703 & 2.8950  \\
        fr3\_walking\_half & \textbf{1.9705} & \textbf{0.8776} & 7.124 & 1.9094 \\
        fr3\_sitting\_static & 0.2934 & \textbf{0.2518} & \textbf{0.2890} & 0.2593  \\
        %     \\\hline
        %        \multicolumn{1}{c||}{mean} & \textbf{0.069} & \textbf{0.060} & 0.205 & 0.186 \\ 
        \hline \hline
    \end{tabular}
    \label{tab.rot}
\end{table}

An extensive experimental validation is performed in terms of absolute trajectory error (ATE), along with relative ose error (RPE) through root mean squared error (RMSE) and median absolute error (MAE) over the entire trajectory. 
As shown in Table \ref{tab.ate}, \ref{tab.rpe}, \ref{tab.rot} the proposed GMC aided system far surpasses the pure vision-based approach under dynamic scenes.
%Fig.~\ref{fig:traj_ni} illustrates the plot of overhead 2-D trajectory from the No. 07 and No. 14 dataset. 
It can be seen that the trajectory estimation from the GMC aided system is much closer to the ground truth with lower drift and lower rate of failure cases.

%\begin{figure}[!t]
%    \centering
%    \includegraphics[width=1.0\linewidth]{./figures/trajectory}
%    \caption{\textbf{Some examples of trajectory estimation with and without UWB aids.} The trajectory estimation with UWB aids is proved to be much closer to the ground truth. }
%    \label{fig:traj_ni}
%\end{figure}

\subsection{Efficiency Evaluation}\label{subsec:efficiency}

\begin{table}[!t]
    \centering
    \vspace*{-3mm}
    \caption{Average runtime of the GMC aided system. `ORB+GMC' means the GMC aided ORBSLAM system.}
    \begin{tabular}{c|c|c|c|c}
        \hline\hline
        Platform & Method & Tracking & Mapping & Total \\\hline 
        \multirow{2}{*}{Laptop} & ORB+GMC & 116.7ms & 9.4ms & 126.1ms \\ [1pt]
        & ORBSLAM  & 7.5ms & 9.2ms & 15.7ms\\\hline 
%        \multirow{2}{*}{Up-Board\textsuperscript{\textregistered} } & ORB+GMC & 67ms & 5ms & 32ms \\[1pt]
%        & ORBSLAM  & 67ms & 9.6ms & 32ms \\
        \hline
    \end{tabular}
    \label{tab:runtime}
\end{table}

We evaluate the efficiency of the proposed system on both tracking and mapping over all datasets. 
The average running time on two platforms is given separately in Table \ref{tab:runtime}. The added time on top of ORBSLAM can be divided into two parts. One part is the GMC time which has complexity $O(N)$. The other part is the additional time for feature extraction. To maintain the property of motion coherence, we have to increase the number of features to be detected.

Note that the proposed method is not optimized due to specifically for SLAM system. We can use GPU parallel architecture and TBB parallel library to accelerate the feature extraction and loop access of matrices. In this way the efficiency of our proposed method can be further improved.

Therefore, the proposed GMC aided approach does not corrupt the instantaneity of the original method.

\section{Conclusions}

We propose GMC, a statistical filter for dynamic objects during pose estimation tasks, by partitioning of different motion patterns based on the  number of neighboring keypoints. A simple and fast system is developed based on the algorithm which can be validated by experiment results.

%\section*{References}
%\bibliographystyle{bibtex/spbasic}
%\bibliography{ref}

\end{document}